\documentclass[runningheads]{llncs}

\usepackage[T1]{fontenc}
\usepackage{amsmath}
\usepackage{hyperref}
\usepackage{multirow}
\usepackage{lipsum}
\usepackage{eqparbox}
\usepackage{graphicx}
\usepackage{tabularx}
\usepackage{subcaption}
\usepackage{appendix}
\usepackage{siunitx}
\usepackage{booktabs}
\usepackage{soul}
\usepackage{setspace}
\usepackage{array,ragged2e}
\usepackage{pifont}

\newcommand{\cmark}{\ding{51}}%
\newcommand{\xmark}{\ding{55}}%

\newcolumntype{C}[1]{>{\centering\arraybackslash}p{#1}}
\newcolumntype{R}[1]{>{\RaggedLeft\arraybackslash}p{#1}}
\usepackage{caption}
\newcommand{\bftab}{\fontseries{b}\selectfont}

\usepackage{color}
\usepackage{xcolor}

\makeatletter
\newcommand*\bigcdot{\mathpalette\bigcdot@{.5}}
\newcommand*\bigcdot@[2]{\mathbin{\vcenter{\hbox{\scalebox{#2}{$\m@th#1\bullet$}}}}}
\makeatother

\begin{document}
\title{\textsc{SoftQE}: Learned Representations of Queries Expanded by LLMs}
%
%
\author{Varad Pimpalkhute\inst{1,}\thanks{Work done as an intern at Amazon} \and
John Heyer\inst{2} \and
Xusen Yin\inst{2} \and
Sameer Gupta \inst{2}}
%
%
\institute{University of Massachusetts Amherst \and
Alexa AI, Amazon\\
\email{pimpalkhutevarad@gmail.com,\{heyjohn,yxusen,gupsam\}@amazon.com}}
\maketitle              
\begin{abstract}
We investigate the integration of Large Language Models (LLMs) into query encoders to improve dense retrieval without increasing latency and cost, by circumventing the dependency on LLMs at inference time. \textsc{SoftQE} incorporates knowledge from LLMs by mapping embeddings of input queries to those of the LLM-expanded queries. While improvements over various strong baselines on in-domain MS-MARCO metrics are marginal, \textsc{SoftQE} improves performance by 2.83 absolute percentage points on average on five out-of-domain BEIR tasks.


\end{abstract}

\section{Introduction}

Query expansion \cite{PRF,rocchio} methods aim to expand search queries with additional terms to improve downstream information retrieval (IR) performance. Expansion terms can come directly from highly ranked documents, as in pseudo relevance feedback based methods like RM3 \cite{PRF,PLM}, or from generative models as in methods like GAR \cite{gar}. While query expansion can mitigate the \textit{token mismatch} problem that plagues sparse retrieval methods like BM25 \cite{bm25}, which depend on token overlap between queries and documents, dense retrieval methods \cite{realm,dpr} offer a natural solution by embedding queries and documents in a shared feature space wherein queries and documents with strong \textit{semantic} overlap are close.

\begin{figure}[htbp]
    \centering
    \includegraphics[width=0.9\linewidth]{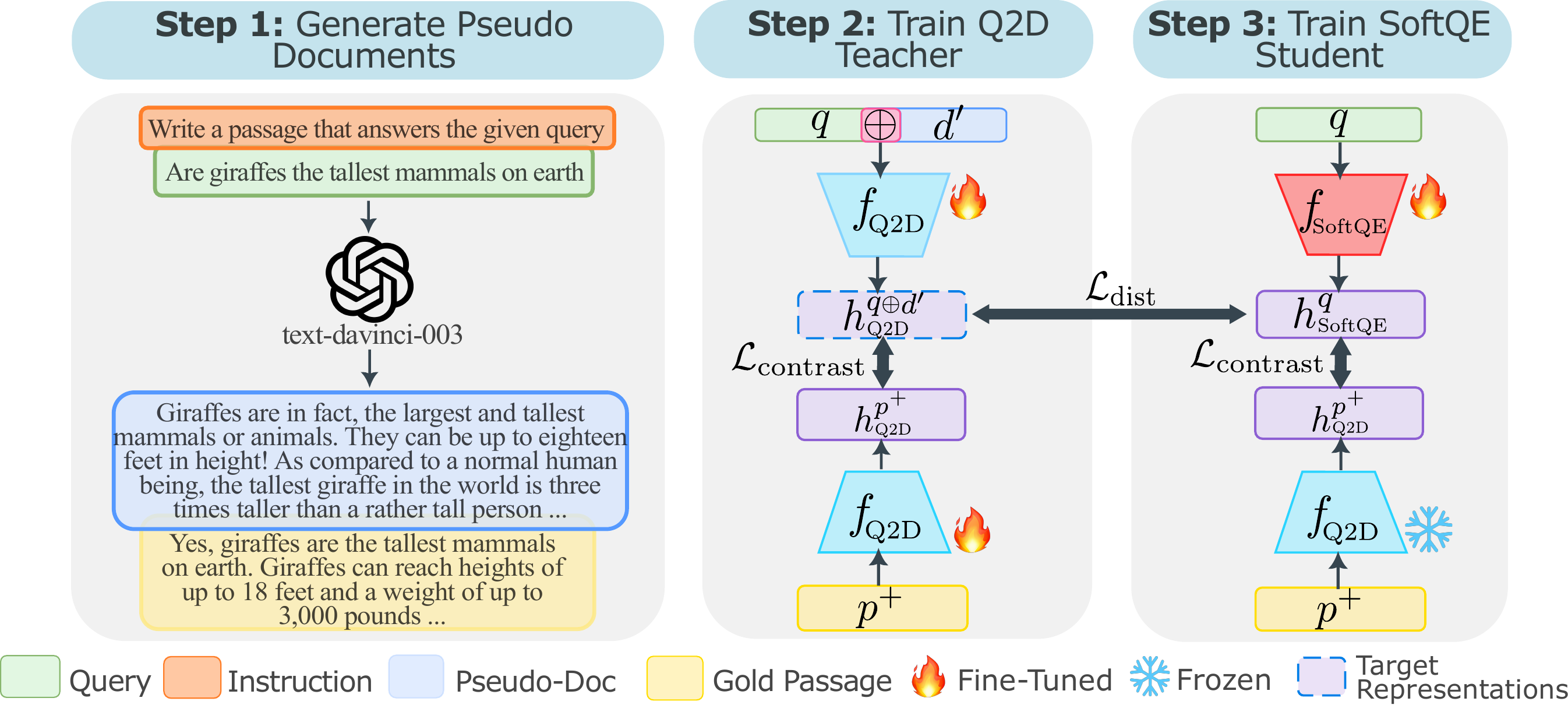}
    \caption{\small Overview of the \textsc{SoftQE} approach. \textbf{Step 1:} Given a query, prompt an LLM to generate a pseudo-document $d'$, as in Q2D~\cite{Q2D}. \textbf{Step 2:} Train teacher encoder using the Q2D method and expanded queries from Step 1 ($q \oplus d'$). \textbf{Step 3:} Train \textsc{SoftQE} encoder to align query representations with the expanded query representations from Step 2, in addition to the standard contrastive objective. $h^x_y$ denotes the \textit{representation} (e.g., the last hidden state of the \texttt{CLS} token) given an input $x$ and encoder $y$.}
    \label{fig:proposed_methodology_2}
\end{figure}

Recent methods \cite{HyDE,QE,Q2D} prompt Large Language Models (LLMs) \cite{gpt3,palm,llama} to expand queries with relevant terms or "pseudo-documents" that resemble real passages from the corpus. Perhaps surprisingly, \textit{query2doc} (Q2D) \cite{Q2D} demonstrates improved performance of \textit{dense} retrievers, indicating that LLM-based query expansion can facilitate learning the semantic overlap between underspecified queries and document corpora. However, adding an LLM to a real-time IR pipeline is often prohibitively expensive in terms of both cost and latency. Motivated by both the promise of LLM-based query expansion for dense retrieval \textit{and} its impracticality, we propose \textit{\textbf{Soft} \textbf{Q}uery \textbf{E}xpansion} (\textsc{SoftQE}), wherein we learn to estimate the representations of LLM expansions \textit{offline} during training, thus circumventing the dependency on LLMs at runtime as shown in Figure \ref{fig:proposed_methodology_2}. \textsc{SoftQE} performs at least as well as baseline dense retrievers such as DPR~\cite{dpr}, and stronger alternatives combining large-scale pretraining and cross-encoder distillation such as SimLM~\cite{simlm} and E5~\cite{e5}, on in-domain MS-MARCO \cite{msmarco}, TREC DL 2019 and 2020 datasets \cite{trecdl19,trecdl20}. Further, \textsc{SoftQE} significantly improves upon these baselines for a majority of out-of-domain BEIR \cite{beir} tasks. Our findings corroborate those of Q2D, specifically that the increase in retrieval performance diminishes when combined with stronger encoders. However, we observe measurable improvements in the zero-shot setting, suggesting that information learned through the \textsc{SoftQE} objective is complementary to other forms of distillation, such as distillation from a cross-encoder.

\vspace{-0.7cm}
\section{Method}
\label{sec:methodology}
\subsubsection{Expanded Queries.} An expanded query $q^+$ is formed by appending a pseudo-document $d'$ to the original query, $q$: 
\begin{equation}
    q^+ = \textit{q } \oplus g_{\phi}(\mathcal{I}, q),
   \label{eq:query_exp}
\end{equation}
where, $g_{\phi}$ is an LLM that generates pseudo document ($d'$) with prompt $\mathcal{I}$, employing techniques such few-shot, chain of thought \cite{cot}, etc. We use the pseudo-documents released\footnote{Pseudo-documents generated using text-davinci-003 for MS MARCO queries are released by \cite{Q2D} here: \href{https://huggingface.co/datasets/intfloat/query2doc_msmarco}{https://huggingface.co/datasets/intfloat/query2doc\_msmarco}} with Q2D \cite{Q2D}, which were generated by \textit{text-davinci-003} \cite{gpt3} using an instruction and examples of positive query/document pairs from MS MARCO \cite{msmarco}. An example pseudo document is shown in Figure \ref{fig:proposed_methodology_2}.

\subsubsection{Dual-Encoder Training.}  Dual encoders are typically trained by optimizing a contrastive objective \cite{dpr}:   
\begin{equation}
    \mathcal{L}_{\text{cont}} = - \log \Big(\frac{e^{h_q \bigcdot h_{p^+}}}{e^{h_q \bigcdot h_{p^+}} + \sum_{i=1}^{N} e^{h_q \bigcdot h_{p^-_i}}} \Big),
    \label{eq:contrast}
\end{equation}
where $h_q$ and $h_p$ represent query and passage embeddings, respectively, and $N$ is the number of negative passages. In Q2D, embeddings of \textit{expanded} query inputs ($h_{q^+}$) are learned, and BM25 hard negatives are used.

\vspace{-0.4cm}
\subsubsection{\textsc{SoftQE} Objective.} Driven by the superior performance of Q2D, we seek to align representations of queries with their expanded counterparts. We do so by introducing an additional distance component\footnote{In practice, we find no significant difference between distance metrics, so we simply use mean squared error (MSE).}, $\mathcal{L}_{\text{dist}}$, to the loss:
\begin{equation}
    \mathcal{L}_{\text{SoftQE}} = \alpha\mathcal{L}_{\text{dist}}(f_{\theta}(q^+),f_{\psi}(q)) + (1-\alpha) \mathcal{L}_{\text{cont}},
    \label{eq:distillation}
\end{equation}
where $f_{\theta}$ and $f_{\psi}$ are transformer-based \cite{transformer} encoders that map expanded queries and queries to vectors in the learned embedding space respectively, and $\alpha$ is a hyper parameter that controls the weight assigned to each component of the loss, as in knowledge distillation \cite{TradKD}. In other words, the expanded query representations produced by the Q2D encoder (teacher) serve as target query representations used to distill information into the \textsc{SoftQE} query encoder (student). Importantly, the feature space is \textit{pre-defined} by the Q2D dual-encoder, rather than updated during training. Accordingly, we only learn to embed \textit{queries}, and reuse the Q2D encoder to produce passage embeddings as they are already well-aligned with the target query representations.

We additionally experiment with state-of-the-art dense retrievers \cite{simlm,e5} that are trained using KL divergence from cross-encoder scores \cite{qu-etal-2021-rocketqa}. We apply \textsc{SoftQE} to distilled retrievers by simply combining the 3 objective terms with an additional weight controlled by $\beta$:
\begin{equation}
    \mathcal{L}_{\text{SQE+KD}} = \alpha\mathcal{L}_{\text{dist}}(f_{\theta}(q^+),f_{\psi}(q)) + (1-\alpha) \left[ \beta\text{KL}(f_{\theta}, f_{\text{CE}}) + (1 - \beta)\mathcal{L}_{\text{cont}} \right],
    \label{eq:simlm_distillation}
\end{equation}
as we find the information distilled through cross-encoder scores and expanded query representations to be complementary.

\vspace{-0.1cm}
\section{Experiments}
\label{sec:results}

\subsubsection{Datasets, Metrics, and Baselines.} For in-domain evaluation, we use the MS MARCO Passage Ranking ~\cite{msmarco}, TREC DL 2019~\cite{trecdl19} and TREC DL 2020~\cite{trecdl20} datasets.
Following Q2D \cite{Q2D}, we evaluate zero-shot performance on five low-resource tasks from the BEIR benchmark~\cite{beir}, namely: SciFact, NFCorpus, Trec-Covid, DBPedia and Touche-2020. Evaluation metrics include $\text{MRR@10}$, $\text{R@50}$, $\text{R@1k}$, and $\text{nDCG@10}$. We benchmark \textsc{SoftQE} against a DPR~\cite{dpr} dense retrieval baseline, and two state-of-the-art dense retrievers: SimLM~\cite{simlm}, and E5~\cite{e5}.

\subsubsection{Hyperparameters.} We follow the hyperparameter settings used in \cite{Q2D}, with a few distinctions. We initialize our DPR models from $\text{BERT}_{\text{base}}$ \cite{bert}, and our \textsc{SoftQE} variants of SimLM \cite{simlm}, and E5 \cite{e5} from their corresponding public checkpoints. When fine-tuning with cross-encoder distillation, $\beta$ is set to $0.2$, following SimLM \cite{simlm}. We set $\alpha$ to $1.0$ for 3 epochs in order to establish an initial alignment with the target expanded query embeddings, then relax $\alpha$ to $0.2$ as well. This choice is further discussed in Section \ref{subsec:ablations}.

\vspace{-0.9cm}
\begin{table}[htbp]
\singlespacing
    \small
    \centering
    \caption{\small Results on in-domain MS MARCO and TREC DL datasets, grouped by retrievers trained with and without distillation from cross-encoders. \underline{Underline}: best result including Q2D, which requires an LLM at inference time; \textbf{Bold}: highest result among non-Q2D solutions; $^{*}$: our reproduction; $\dagger$: denotes statistical significance with a p-value less than 0.05 using a paired T-test.} 
    \begin{tabular*}{0.85\textwidth}{@{\extracolsep{\fill}}p{0.15\textwidth} *{5}{c}}
        \hline
        \multirow{2}{*}{\bftab{Method}} & \multicolumn{3}{c}{\bftab{MS MARCO Dev Set}} & \bftab{TREC DL 19} & \bftab{TREC DL 20} \\
        \cline{2-4}
        & MRR@10 & R@50 & R@1k & nDCG@10 & nDCG@10 \\
        \hline
        \multicolumn{6}{l}{\textit{Dual-encoder without distillation}} \\
        $\text{DPR}^*$ & 33.74 & 80.90 & 96.18 & 64.04 & 62.81 \\
        $\text{+ \textsc{SoftQE}}$ & \bftab{33.87} & \bftab{ 81.24}$^\dagger$ & \bftab{ 96.25}$^\dagger$ & \bftab{ 65.22}$^\dagger$ & \bftab{ 63.80}$^\dagger$ \\
        $\text{+ Q2D}^*$ & \underline{35.26} & \underline{82.78} & \underline{97.21}& \underline{70.54} & \underline{66.68} \\
        \hline
        \multicolumn{6}{l}{\textit{Dual-encoders distilled from cross-encoders}} \\
        $\text{SimLM}^*$ & 41.13 & 87.78 & \bftab{98.69} & 71.40 & 69.68 \\
        $\text{+ \textsc{SoftQE}}$ & \bftab{41.15} & \bftab{ 87.93}$^\dagger$ & { 98.61}$^\dagger$ & { 70.50}$^\dagger$ & { 70.10}$^\dagger$   \\
        $\text{+ Q2D}^*$ & \underline{41.45} & \underline{88.43} & \underline{98.82} & 74.59 & 71.37 \\
        $\text{E5}^*$ & 40.70 & 87.13 & 98.50 & 72.52 & 71.38 \\
        $\text{+ \textsc{SoftQE}}$ & { 40.30}$^\dagger$ & 87.22 & 98.50 & \bftab{ 72.82}$^\dagger$ & \bftab{ 71.73}$^\dagger$ \\
        $\text{+ Q2D}^*$ & 40.93 & 87.95 & 98.76 & \underline{75.03} & \underline{73.27} \\
        \hline

    \end{tabular*}

    \small \label{tab:in_domain}
\end{table}

\vspace{-0.9cm}

\subsubsection{Results.} We first evaluate the performance on in-domain datasets (Table \ref{tab:in_domain}). \textsc{SoftQE} consistently improves upon DPR across all metrics on MS MARCO, TREC DL 19 and TREC DL 20 datasets. When evaluating the performance against dual-encoders distilled from cross-encoders, we notice that \textsc{SoftQE} and SimLM perform closely with \textsc{SoftQE} slightly underperforming in $\text{R@1k}$ on MS MARCO and $\text{nDCG@10}$ on TREC DL2019. Similarly, \textsc{SoftQE} results in marginal improvements over E5. This finding corroborates the claim in \cite{Q2D} that improvements diminish when encoders are distilled from strong cross-encoders.

\vspace{-0.5cm}
\begin{table}[htbp]
    \centering
    \small
    \caption{Results on out-of-domain BEIR benchmark datasets by nDCG@10. \underline{Underline}: best result including Q2D, which requires an LLM at inference time; \textbf{Bold}: highest result among non-Q2D solutions; $^{*}$: our reproduction\protect\footnotemark; $\dagger$: denotes statistical significance.}
    \begin{tabular*}{0.9\textwidth}{@{\extracolsep{\fill}}p{0.18\textwidth} *{5}{c}|c}
        \hline
        \bftab{Method} & SciFact & NFCorpus & Trec-Covid & DBPedia & Touche-2020 & Average\\
        \hline
        $\text{DPR}^*$ & \bftab{51.85} & 25.72 & 44.81 & 30.99 & 19.91  & 34.65\\
        $\text{+ \textsc{SoftQE}}$ & 49.81$^{\dagger}$ & \bftab{ 25.73} & \bftab 60.02$^\dagger$ & \bftab{31.82}$^\dagger$ & \bftab{20.59} & \textbf{37.59}\\
        \hline
        $\text{SimLM}^*$ & 61.42 & \underline{\bftab{32.38}} & 52.90 & 35.06 & 19.21  & 40.19\\
        $\text{+ \textsc{SoftQE}}$ & \underline{\bftab{61.72}} & 32.34 & \underline{\bftab{61.78}} & \bftab{36.75} & \bftab{21.94}  & \bftab 42.91\\
        $\text{+ Q2D}$~\cite{Q2D} & 59.50 & 32.10 & 59.90 & \underline{38.30} & \underline{25.60} & \underline{43.08} \\

        \hline
    \end{tabular*}
    \label{tab:out_domain}
\end{table}

\footnotetext{We could not reliably reproduce the E5 results on BEIR datasets, but Q2D did not yield significant improvements when applied to E5, so we assume the same for \textsc{SoftQE} and omit E5 from zero-shot evaluation.}

Table \ref{tab:out_domain} highlights the zero-shot evaluation results on out-of-domain datasets from BEIR. \textsc{SoftQE} considerably outperforms DPR and SimLM, by $2.94$ and $2.72$ absolute percentage points, respectively, averaged across tasks. \textsc{SoftQE} yields marginal differences in performance on tasks where Q2D results in regressions (SciFact and NFCorpus), but substantial improvements on the remaining tasks when applied to either DPR or SimLM, indicating that \textsc{SoftQE} is complementary to cross-encoder distillation.

\section{Discussion}
\label{subsec:ablations}

\noindent\textbf{Is Fine-tuning on Expanded Queries Necessary?} Traditional query expansion methods applied to lexical systems do not require modifications to the retrieval algorithm. Q2D~\cite{Q2D}, however, requires fine-tuning the dense retriever on expanded queries, as demonstrated by the the difference between the first 2 rows in Table \ref{tab:ablations_2}. Simply passing expanded queries to an off-the-shelf DPR model actually deteriorates performance, which is somewhat surprising given the model's ability to effectively embed queries and passages \textit{independently}.

\vspace{-0.5cm}
\begin{figure}
  \begin{minipage}{0.46\textwidth}
    \centering
    \small
    \captionof{table}{MS Marco MRR@10 of DPR and Q2D with query ($q$) and expanded query ($q^+$) inputs. DPR has not been trained with expanded query inputs, while Q2D has.}
    \begin{tabular}{lcc}
      \hline
      \textbf{Method} & \textsc{Trec dl19} & \textsc{Trec dl20} \\
      \hline
    DPR($q^+$) & 61.65 & 59.45 \\
    + Q2D($q^+$) & \textbf{70.54} & \bftab{66.68} \\
      \hline
      DPR($q$) & 64.04 & 62.81 \\
    + Q2D($q$) & 57.78 & 57.12 \\
      \hline
    \end{tabular}
    \label{tab:ablations_2}
  \end{minipage}
  \hfill 
  \begin{minipage}{0.52\textwidth}
    \centering
    \small
    \captionof{table}{TREC $\text{nDCG@10}$ across four variations of $\alpha$ in the training objective: ($\alpha\mathcal{L}_{\text{cont}}$+(1-$\alpha)\mathcal{L}_{\text{dist}}$). In "Warm up", we set alpha to 1 for the first 3 epochs, then to $0.2$ for the remaining 3.}
    \begin{tabular*}{\textwidth}{p{1.76cm}ccc}
      \hline
      \textbf{Method} & $\alpha$ & \textsc{Trec dl19} & \textsc{Trec dl20} \\
      \hline
      $\mathcal{L}_{\text{dist}}$ Only & 1 & 61.62 & 62.81\\
      $\mathcal{L}_{\text{cont}}$ Only & 0 & 64.66 & 63.42\\
      Combined & 0.2 & 63.33 & 63.03\\
      Warm up & 1$\rightarrow$0.2  & \bftab 65.23 & \bftab 63.79\\
      \hline
    \end{tabular*}
    \label{tab:ablations_3}
  \end{minipage}
\end{figure}
\vspace{-0.5cm}

\begin{figure}[ht]
  \begin{minipage}{0.48\textwidth}
    \centering
    \includegraphics[width=\linewidth]{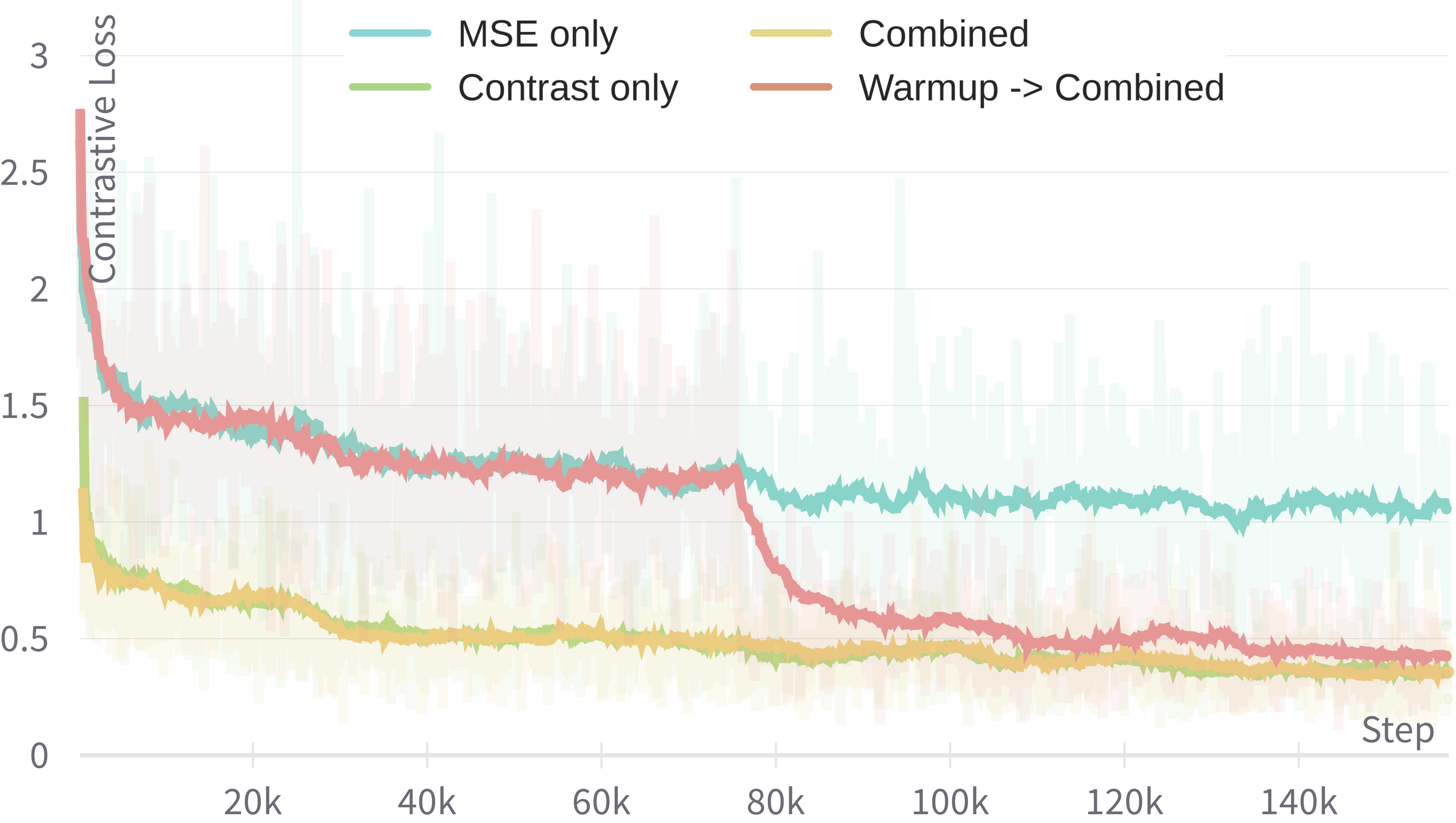}
  \end{minipage}%
  \hfill
  \begin{minipage}{0.48 \textwidth}
    \centering
    \includegraphics[width=\linewidth]{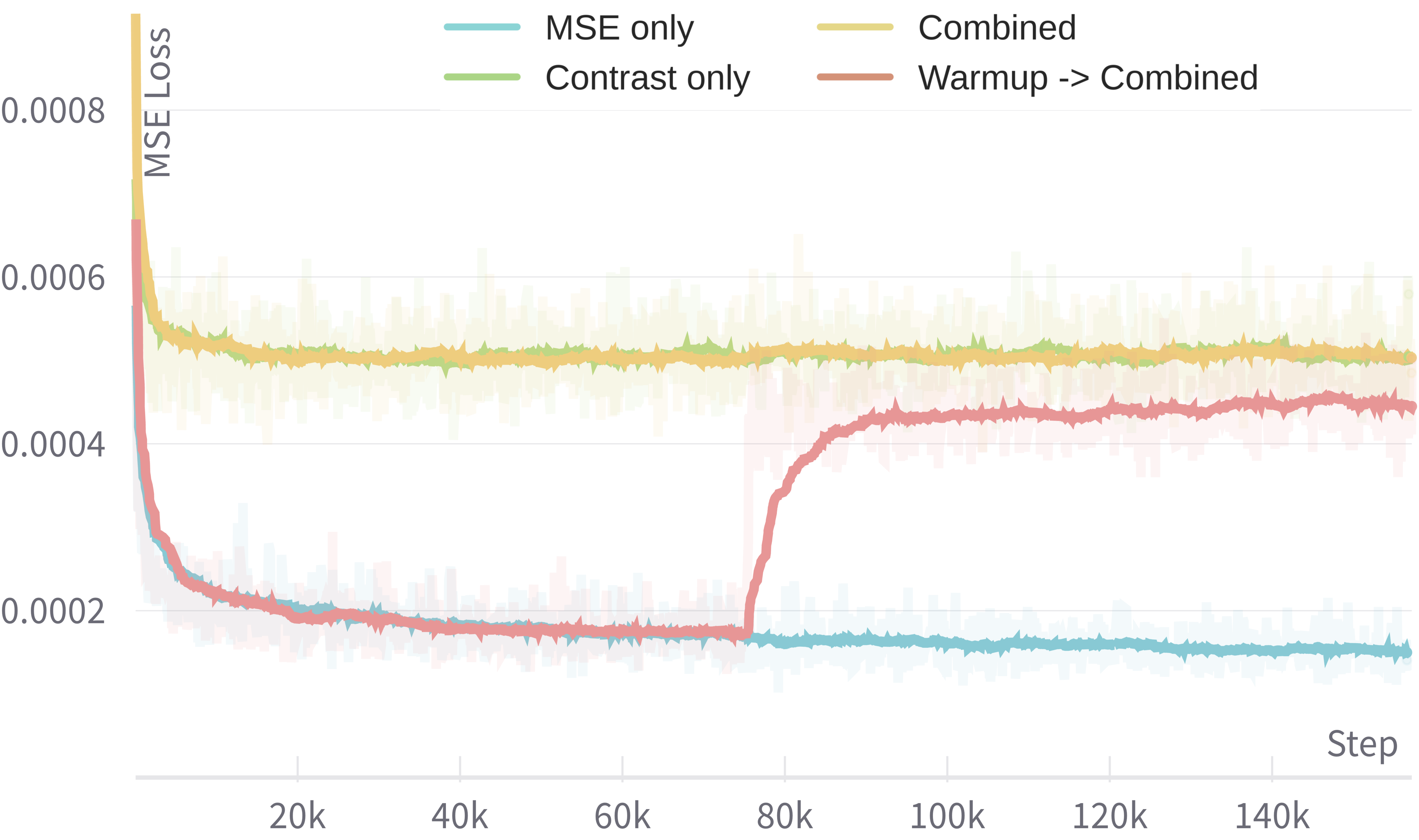}
  \end{minipage}%
  \caption{\small Training curves of four settings of $\alpha$ shown in Table \ref{tab:ablations_3}. \textbf{Left}: Contrastive Loss - \textit{does it reject negative documents?}  \texttt{MSE-only} performs the worst in terms of contrastive loss, while \texttt{Warmup$\rightarrow$Combined} converges to the same loss as \textit{Combined}. \textbf{Right}: MSE Loss - \textit{is it close to the teacher?} \texttt{Contrast-only} has the highest MSE loss, while \texttt{Warmup$\rightarrow$Combined} MSE loss increases after the warmup, but converges to a value noticeably lower than \texttt{Combined}.}
  \label{fig:loss_plots}
\end{figure}

\noindent\textbf{Combining $L_{\text{dist}}$ and $L_{\text{cont}}$.}
In Table \ref{tab:ablations_3}, we explore four variations of the training objective in Equation \ref{eq:distillation} to determine how to balance supervision from labeled passages vs. target representations. A perfect mapping ($L_{\text{dist}}=0$) between query and expanded query representations would yield Q2D performance, but is not realistic, as evident by the subpar performance of "$L_{\text{dist}}$ Only". Combining the two losses by setting $\alpha$ to $0.2$ results in query embeddings that are no closer to the target embeddings produced when using only a contrastive loss, as shown by the MSE Loss plot in Figure \ref{fig:loss_plots} (right). To remedy this, we propose a step-wise "warm up" method, in which we set $\alpha$ to 1 ($L_{\text{dist}}$ only) for 3 epochs to establish a strong alignment with the target representations, then relax $\alpha$ to $0.2$ for the remaining 3 epochs. Figure \ref{fig:loss_plots} demonstrates that this reduces $L_{\text{dist}}$ while negligibly impacting $L_{\text{cont}}$, resulting in the best performance in Table \ref{tab:ablations_3}.

\noindent \textbf{Should we also Fine-tune the Passage Encoder?} We decided not to fine-tune the passage encoder during \textsc{SoftQE} training, because it allowed us to re-use the Q2D passage-encoder. Intuitively, this means that the space in which passages and queries are embedded is the same as in Q2D. In Table \ref{tab:ablations_4} we show that, on average, fine-tuning the passage encoder results in reduced performance. This is assuring -- if the passage representations were to change, our Q2D representation targets would be unfounded, as they would no longer be optimally aligned with the passages.

\vspace{-0.5cm}
\begin{figure}
  \begin{minipage}{0.45\textwidth}
    \centering
    \small
    \captionof{table}{Unfreezing the passage encoder during training results in a degradation of performance on TREC nDCG@10.}
    \begin{tabular}{p{1.5cm}p{1.3cm}cc}
      \hline
      \textbf{Freeze Encoder} & \multirow{2}{*}{\textbf{Method}} & \multirow{2}{*}{\textsc{DL19}} & \multirow{2}{*}{\textsc{DL20}} \\
      \hline
     \xmark & \textsc{SoftQE} & \bftab{65.59} & 62.06\\
      \cmark & \textsc{SoftQE} & 65.22 & \bftab{63.80}\\
      \hline
    \end{tabular}
    \label{tab:ablations_4}
  \end{minipage}
  \hfill 
  \begin{minipage}{0.46\textwidth}
    \centering
    \small
    \captionof{table}{Comparing our method to traditional knowledge distillation (using only model predictions) on TREC nDCG@10.}
    \begin{tabular*}{\textwidth}{p{3cm}p{1cm}p{1cm}}
      \hline
      \textbf{Method} & \textsc{DL19} & \textsc{DL20} \\
      \hline
      DPR & 64.04 & 62.81\\
      + \text{Traditional KD} & 61.12 & 62.14\\
      + \textsc{SoftQE}  & \bftab{65.22} & \bftab{63.79}\\
      \hline
    \end{tabular*}
    \label{tab:ablations_5}
  \end{minipage}
\end{figure}
\vspace{-0.5cm}

\noindent \textbf{\textsc{SoftQE} vs. Traditional Knowledge Distillation.} Our method distills the high-dimensional \textit{representation} of the teacher model, as opposed to teacher's \textit{predictions}, as in traditional knowledge distillation. In Table \ref{tab:ablations_5}, we compare our method to a variant in which we compute MSE \textit{only over the scalar-valued scores} produced by the Q2D teacher as our distillation loss. This results in reduced performance, as the score-only distillation model underperforms the DPR baseline, indicating that the teacher's predictions alone do not provide sufficient supervision for estimating the nuanced information contained in the high-dimensional expanded query representations.

\section{Related Work}

\textbf{Document Expansion.} Doc2Query \cite{d2q} attempts to resolve vocabulary mismatch by expanding \textit{documents} with natural language queries whose answers are likely to exist within the document. Document expansion is advantageous because it can be conducted entirely offline during indexing and combined with learned sparse retrieval methods \cite{splade,unicoil,di} to leverage both neural supervision and efficient inverted index algorithms. However, document expansion techniques can significantly increase the size of the index, and must be applied to the entire corpus each time the expansion method is changed, which might be too costly for corpora containing billions of documents.

\noindent \textbf{Knowledge Distillation.} Knowledge distillation \cite{TradKD} methods use the predictions of large \textit{teacher} models to improve the performance of smaller, more practical \textit{student} models. Knowledge distillation is ubiquitous, and has been used to improve dense dual-encoder retrievers via distillation from a cross-encoder \cite{qu-etal-2021-rocketqa}. \textsc{SoftQE} is a form of \textit{indirect} knowledge distillation, wherein a student encoder targets the continuous representations of an architecturally-equivalent teacher model whose discrete, natural language \textit{inputs} have been augmented by an LLM. Recent methods such as Alpaca \cite{alpaca} and Vicuna \cite{vicuna2023} use generations of superior LLMs to improve the performance of smaller, instruction-following models, but these methods imitate "teacher" LLMs by using their outputs as training data directly.

\vspace{-0.25cm}
\section{Conclusion}
\vspace{-0.25cm}
We present \textsc{SoftQE}, a technique to align query-encoder representations with the representations of queries expanded by LLMs. Empirical evaluations demonstrate improvements across several retrieval benchmarks and models, and suggest that \textsc{SoftQE} improves generalization to new domains, as made evident by zero-shot evaluations on BEIR datasets. Importantly, this improvement comes without increasing the cost or latency of dense retrieval at runtime compared to other single vector dual-encoder methods, because an LLM is not required at time of inference. Future work might consider improved prompting strategies, or applying LLM-based supervision to higher-capacity retrieval methods like ColBERTv2~\cite{colbertv2}. To the best of our knowledge, \textsc{SoftQE} is the first attempt to distill strong representations through natural language generation, and we hope that this will inspire efficient solutions to new tasks in the future.

%
%
%
\bibliographystyle{splncs04}
\bibliography{biblio}

\appendix

\section{Additional BEIR Results}

\vspace{-0.5cm}
\begin{table}[htbp]
  \caption{nDCG@10 for the remaining BEIR tasks; $^{*}$: our reproduction (not tested for significance). Q2D was not evaluated on these datasets.}
  \label{tab:beir}
  \centering
  \begin{subtable}{\textwidth} 
    \centering
    \small
    \begin{tabular*}{\linewidth}{@{\extracolsep{\fill}}p{0.18\textwidth} *{7}{c}}
        \hline
        \textbf{Method} & Signal 1m & Trec-News & Quora & NQ & Fiqa & Arguana \\
        \hline
        $\text{DPR}^*$ & \bftab{24.15} & 35.21 & 84.05 & 43.72 & 24.47  & 28.80\\
        $\text{+ \textsc{SoftQE}}$ & 22.18 & \bftab{36.43} & \bftab{84.32} & \bftab{44.19} & \bftab{25.07} & \bftab{31.11}\\
        \hline
    \end{tabular*}
    \label{subtab:table1}
  \end{subtable}
  
  \begin{subtable}{\textwidth} 
    \centering
    \small
    \begin{tabular*}{\linewidth}{@{\extracolsep{\fill}}p{0.18\textwidth} *{6}{c}}
        \hline
        \textbf{Method} & Scidocs & BioASQ & HotpotQA & Climate Fever & Fever  & Avg.\\
        \hline
        $\text{DPR}^*$ & \bftab{11.79} & 25.09 & \bftab{49.58} & 17.20 & 65.36 & 37.22 \\
        $\text{+ \textsc{SoftQE}}$ & 11.35 & \bftab{27.05} & {48.36} & \bftab{18.40} & \bftab{67.38} & 37.80 \\
        \hline
    \end{tabular*}
    \label{subtab:table2}
  \end{subtable}
\end{table}

\end{document}